\documentclass{article}

\usepackage{PRIMEarxiv}

\usepackage[utf8]{inputenc} % allow utf-8 input
\usepackage[T1]{fontenc}    % use 8-bit T1 fonts
\usepackage{hyperref}       % hyperlinks
\usepackage{url}            % simple URL typesetting
\usepackage{booktabs}       % professional-quality tables
\usepackage{amsfonts}       % blackboard math symbols
\usepackage{nicefrac}       % compact symbols for 1/2, etc.
\usepackage{microtype}      % microtypography
\usepackage{lipsum}

\usepackage{amsmath}
\usepackage{cleveref} % for \Cref and \cref cross-referencing

\DeclareMathOperator{\argsort}{argsort}
\usepackage{fancyhdr}       % header
\usepackage{graphicx}       % graphics
\graphicspath{{media/}}     % organize your images and other figures under media/ folder

\usepackage{algorithmic}
\usepackage{float}
\usepackage{bm}
\usepackage{graphicx}
\usepackage{graphics}
\usepackage{caption}
\usepackage{subcaption} 
\usepackage[ruled, vlined]{algorithm2e}
\usepackage{comment} 

\title{TrustFed: Enabling trustworthy medical AI under data privacy constraints
}

\author{
  Vagish Kumar \\
  Department of Applied Mechanics \\
  Indian Institute of Technology Delhi \\
  \texttt{vagish.kumar@am.iitd.ac.in} \\
    \And
  Syed Bahauddin Alam \\
  Grainger College of Engineering \\
  Nuclear, Plasma \& Radiological Engineering Department \\
  National Center for Supercomputing Applications, Urbana, IL, USA \\
  \texttt{alams@illinois.edu}\\
  \And
  Souvik Chakraborty \\
  Department of Applied Mechanics \\
  Yardi School of Artificial Intelligence (ScAI) \\
  Grainger College of Engineering \\
  Indian Institute of Technology Delhi \\
  University of Illinois Urbana-Champaign, Urbana, IL, USA \\
   \texttt{souvik@am.iitd.ac.in} \\
}

\begin{document}
\maketitle

\begin{abstract}
Protecting patient privacy remains a fundamental barrier to scaling machine learning across healthcare institutions, where centralizing sensitive data is often infeasible due to ethical, legal, and regulatory constraints. Federated learning has emerged as a viable alternative by enabling privacy-preserving, multi-institutional model training without sharing raw patient data; however, in realistic clinical settings, severe data heterogeneity, site-specific biases, and class imbalance undermine predictive reliability and render existing uncertainty quantification approaches ineffective. Here, we present TrustFed, the first federated uncertainty quantification framework that provides distribution-free, finite-sample coverage guarantees under heterogeneous and imbalanced healthcare data, without requiring access to centralized raw patient data. TrustFed introduces a representation-aware client assignment mechanism that leverages the model's internal representations to enable effective calibration across heterogeneous institutions, together with a soft-nearest threshold aggregation strategy that mitigates assignment uncertainty while producing compact and reliable prediction sets. Using over 430,000 medical images across six clinically distinct imaging modalities, we conduct one of the most comprehensive evaluations of uncertainty-aware federated learning in medical imaging to date, demonstrating robust coverage guarantees across datasets with diverse class cardinalities and imbalance regimes. By validating TrustFed at this scale and clinical breadth, our study moves uncertainty-aware federated learning from proof-of-concept demonstrations toward clinically meaningful, modality-agnostic deployment, positioning statistically guaranteed uncertainty as a core requirement for next-generation healthcare AI systems. We anticipate our study to: (1) enable uncertainty-aware federated learning for safety-critical clinical workflows where calibrated confidence is essential for decision-making, (2) establish representation-aware conformal prediction as a practical foundation for trustworthy multi-institutional healthcare AI, and (3) catalyze a paradigm shift toward statistically guaranteed, privacy-preserving collaboration across healthcare systems, alleviating the need for centralized data sharing while improving reliability and equity in medical machine learning.
\end{abstract}


\section{Introduction}
\label{sec:introduction}
The rapid digitization of healthcare has led to an unprecedented accumulation of medical data across hospitals, diagnostic laboratories, and imaging centers worldwide \cite{world2025global, organisation2019health}. While this data holds immense potential for improving diagnosis, prognosis, and clinical decision-making through machine learning (ML), its use is fundamentally constrained by privacy, ownership, and regulatory barriers \cite{price2019privacy, rieke2020future}. Strict legal frameworks, including the Health Insurance Portability and Accountability Act (HIPAA) in the United States \cite{edemekong2018health}, the General Data Protection Regulation (GDPR) in Europe \cite{regulation2018general}, and the Digital Personal Data Protection Act (DPDP) in India \cite{gupta2024digital}, severely limit the centralization and cross-border sharing of patient data. These constraints are not merely administrative obstacles but reflect ethical imperatives central to patient trust and clinical responsibility. As a result, many healthcare ML models remain trained on siloed, institution-specific datasets, limiting their generalizability, robustness, and clinical reliability. Federated learning (FL) \cite{mcmahan2017communication, li2020federated, reddi2020adaptive, acar2021federated} has emerged as a compelling and promising paradigm to overcome these barriers by enabling collaborative model training without transferring raw patient data outside institutional  \cite{kairouz2021advances, bonawitz2022federated}. By exchanging only model updates rather than sensitive data, FL ensures privacy-preserving, multi-institutional learning at scale, and has shown encouraging performance across a range of medical imaging and clinical prediction tasks \cite{sheller2020federated, kaissis2020secure, dayan2021federated, pati2022federated}. However, real-world healthcare data are inherently heterogeneous: institutions differ in acquisition protocols, patient demographics, disease prevalence, and annotation practices \cite{kelly2019key, zhao2018federated, liu2022real, li2022federated}. Such heterogeneity, often compounded by extreme class imbalance \cite{johnson2019survey}, challenges a core assumption underlying most ML models, that training and deployment data are identically distributed \cite{de1980partial}. Consequently, even when FL achieves strong average predictive accuracy, its predictions may be unreliable or poorly calibrated for specific patients or sites \cite{guo2017calibration, ovadia2019can}, undermining trust in safety-critical clinical workflows.
Reliable deployment of medical AI systems requires not only high predictive accuracy but also principled and transparent quantification of uncertainty \cite{kompa2021second, marconi2025show, loftus2022uncertainty}. In clinical practice, understanding when a model is uncertain is often as critical as understanding what it predicts, particularly in safety-critical workflows such as diagnosis, triage, treatment planning, and human-AI collaboration \cite{esteva2019guide}. Uncertainty estimates directly influence clinical trust, determine when human oversight is required, and shape downstream decision-making under risk. However, despite rapid progress in federated learning for medical imaging, uncertainty quantification remains markedly underdeveloped in federated medical AI \cite{antunes2022federated}. While methods such as Bayesian approximations \cite{kendall2017uncertainties} and deep ensembles \cite{lakshminarayanan2017simple} have advanced uncertainty estimation in centralized settings, their extension to federated environments with heterogeneous data remains largely unaddressed.
Existing approaches largely rely on heuristic confidence scores, post-hoc calibration, or centralized validation procedures that implicitly assume data exchangeability \cite{vovk2005algorithmic} and homogeneous data distributions. These assumptions are routinely violated in federated healthcare settings, where data are distributed across institutions with distinct acquisition protocols, patient populations, disease prevalence, and annotation practices. As a consequence, uncertainty estimates produced by such methods may be misleading, either overly conservative, resulting in impractically large abstention regions, or falsely confident, masking model failure modes in out-of-distribution clinical settings. Crucially, the inability to provide reliable uncertainty under privacy constraints undermines the very promise of federated learning as a trustworthy alternative to centralized training. Addressing uncertainty in federated medical AI is therefore not a secondary refinement, but a foundational requirement for clinical deployment, regulatory acceptance, and meaningful human-AI collaboration \cite{begoli2019need}.
In this work, we argue that trustworthy federated medical AI demands a fundamental rethinking of uncertainty quantification under data heterogeneity and privacy constraints. We introduce TrustFed, a sample-adaptive federated conformal prediction framework that provides statistically valid uncertainty estimates without requiring access to centralized patient data. Unlike prior federated conformal prediction methods that rely on global pooling or fixed calibration strategies \cite{lu2023federated, plassier2023conformal}, TrustFed adapts uncertainty estimates at test time by leveraging representation-aware client assignment and soft-nearest threshold aggregation. This design explicitly accounts for inter-institutional heterogeneity while preserving privacy, enabling uncertainty estimates that are both reliable and efficient at the level of individual patients. By unifying privacy-preserving learning with statistically guaranteed uncertainty, TrustFed advances federated learning from a mechanism for data access to a framework for clinical trust. The key scientific contributions of this study include:
\begin{itemize}
    \item \textbf{Representation-aware client assignment}: TrustFed introduces a privacy-preserving mechanism to group clients based on model representation similarity without requiring access to raw data or introducing extra communication overhead. This enables more effective calibration in heterogeneous federated settings.
    \item \textbf{Soft-nearest threshold aggregation}: To mitigate the adverse effects of imperfect client assignments, we propose a soft aggregation strategy that smooths the influence of assignment errors. This ensures valid coverage while reducing the average size of prediction sets.
    \item \textbf{Privacy-preserving uncertainty quantification}: The overall framework allows maintaining all calibration and nonconformity computations locally at each client, avoiding any pooling of sensitive information while achieving statistically guaranteed coverage.
    \item \textbf{Comprehensive evaluation across diverse clinical tasks}: The study demonstrates robust, calibrated prediction sets across datasets with heterogeneous class structures, severe imbalance, and multiple imaging modalities, establishing TrustFed as a modality-agnostic approach suitable for real-world deployment.
\end{itemize}

For the first time, this study establishes a federated framework for sample-adaptive, statistically guaranteed uncertainty quantification under inter-institutional heterogeneity and class imbalance. By integrating representation-conditioned calibration with decentralized conformal inference, TrustFed restores finite-sample coverage without centralized pooling of calibration data. Evaluation across six clinically distinct imaging modalities with diverse class structures demonstrates consistent coverage and efficiency under realistic federated conditions.

\section{Method}
\label{sec:Method}
\subsection*{Federated learning setup and limitations of pooled conformal prediction}
We consider a standard federated learning setup consisting of a central server and $K$ clients. Each client $k \in \{1,\dots,K\}$ holds a private dataset $\mathcal{D}_k$ that is never shared with other clients or the server. Let $\{\mathcal{D}_1,\dots,\mathcal{D}_K\}$ denote the data distributed across clients. The global model is a parametric predictor $f(x;w)$ with parameters $w \in \mathbb{R}^d$, trained to minimize the average client-wise loss
\begin{equation}
\min_{w} \; \mathcal{L}(w) \;=\; \frac{1}{K}\sum_{k=1}^{K} {l}_k(w),
\label{eq:global_objective}
\end{equation}
where $l_k(w)$ denotes the empirical loss on client $k$. Training proceeds over synchronous communication rounds $t=1,\dots,T$. In each round, the server broadcasts the current global model $w^{(t)}$, each client performs local optimization for $E$ epochs using mini-batches of size $B$, and the updated local models are aggregated using sample-proportional weights:
\begin{equation}
w^{(t+1)} \;=\; \sum_{k=1}^K \frac{n_k}{n}\, w_k^{(t+1)},
\label{eq:aggregation}
\end{equation}
where $n_k$ is the number of training samples at client $k$ and $n=\sum_k n_k$. To enable uncertainty quantification, each client further partitions its local data into a training set and a calibration set,
\begin{equation}
\mathcal{D}_k = \mathcal{D}_k^{\mathrm{train}} \;\cup\; \mathcal{D}_k^{\mathrm{cal}},
\label{eq:data_splits}
\end{equation}
where $\mathcal{D}_k^{\mathrm{cal}}=\{(x_i^{(k)},y_i^{(k)})\}_{i=1}^{m_k}$ is used for conformal calibration \cite{angelopoulos2023conformal}. {The validity of conformal calibration at each client relies on the assumption that the calibration samples and any future test sample originating from a similar distribution are exchangeable. In the standard centralized setting, this is ensured by random splitting of an i.i.d.\ dataset. In our federated setting, we assume that within each client, the calibration split is drawn exchangeably from the client's local data distribution. Across clients, however, full exchangeability might not hold due to inter-institutional heterogeneity. TrustFed addresses this by not pooling calibration scores globally, but instead selecting, for each test sample, the clients whose learned representations are most similar, thereby constructing a localized calibration neighborhood within which approximate exchangeability is more plausible.}
We focus on multiclass classification with label space $\mathcal{Y}=\{1,\dots,C\}$. Given the trained global model, the nonconformity score is defined as
\begin{equation}
    s(x,y)=1-p_y(x),
    \label{eq:score}
\end{equation}
where $p_y(x)$ is the predicted probability for class $y$. 
For a client-specific calibration set of size $m_k$, let
$\{s_i^{(k)}\}_{i=1}^{m_k}$ denote the corresponding nonconformity scores,
sorted in ascending order. The conformal threshold for client $k$ is
defined using the empirical quantile as
\begin{equation}
\tau_k = s^{(k)}_{\lceil (1-\alpha)(m_k + 1) \rceil},
\label{eq:empirical_quantile}
\end{equation}
Existing approaches pool the calibration scores from all clients,
\begin{equation}
    S=\bigcup_{k=1}^{K} S_k, \quad S_k=\{s(x_i^{(k)},y_i^{(k)})\}_{i=1}^{m_k},
    \label{eq:pooled_scores}
\end{equation}
and compute a global threshold using a $(1-\alpha)$ quantile of the pooled set. While this strategy is simple, it suffers from two fundamental limitations in heterogeneous federated settings. First, pooling implicitly assumes full exchangeability across clients, an assumption violated when client data distributions differ due to class imbalance, acquisition effects, or population shift. As a result, the pooled threshold can be overly conservative or fail to provide valid coverage for test samples originating from specific clients. Second, aggregating calibration scores, even in anonymized form, introduces privacy and information leakage concerns, as score distributions may reveal client-specific characteristics.
These limitations motivate the need for conformal mechanisms that (i) respect partial exchangeability across heterogeneous clients, (ii) avoid centralized pooling of calibration scores, and (iii) adapt prediction thresholds to the representation-level proximity between test samples and client data. In the following subsections, we introduce representation-aware client assignment and soft-nearest threshold aggregation to address these challenges.

\subsection*{Representation-aware client assignment}
In federated learning with heterogeneous client distributions, a single global nonconformity threshold may either be too conservative for some test samples or insufficient for others. This limitation arises because client-specific calibration distributions differ due to variations in acquisition protocols, class prevalence, or patient demographics. To address this, we propose a \emph{representation-aware client assignment} strategy that leverages the global model's learned feature embeddings to identify the clients most relevant to a given test sample, thereby producing more efficient and tailored conformal prediction sets.

Let $\phi(x_i) \in \mathbb{R}^d$ denote the feature embedding of input $x_i$ produced by the global model. Each client $k$ constructs a \emph{calibration feature bank} from its local calibration set:
\begin{equation}
    \mathcal{B}_k = \{ b_{k,1}, b_{k,2}, \dots, b_{k,m_k} \} \subset \mathbb{R}^d,
    \qquad b_{k,j} = \phi(x_j^k),
    \label{eq:feature_bank}
\end{equation}
where $x_j^k$ are the calibration samples for client $k$ and $m_k$ is the number of calibration points. For a test embedding $f_i = \phi(x_{\text{test}})$, the minimum distance to client $k$ is computed as:
\begin{equation}
    d_{i,k} = \min_{1 \le j \le m_k} \lVert f_i - b_{k,j} \rVert_2.
    \label{eq:feature_min_dist}
\end{equation}
Each client computes $d_{i,k}$ locally and returns only this scalar distance to the central server. The server then selects the top-$N_k$ nearest clients in embedding space for the given test sample. This induces a \emph{localized calibration pool} tailored to the test input, which improves the relevance of threshold computation and ensures tighter prediction sets without compromising coverage guarantees. Notably, this procedure preserves privacy since no raw images, labels, or embeddings leave the client; only a scalar distance per test sample is shared.

To highlight the advantages of learned representations, we also evaluated a raw-pixel-based client assignment. In this alternative, each image $X \in \mathbb{R}^{H \times W \times C}$ is vectorized as
\begin{equation}
    r(X) = \text{vec}(X) \in \mathbb{R}^{HWC},
    \label{eq:pixel_repr}
\end{equation}
and each client constructs a \emph{pixel bank} from its calibration images:
\begin{equation}
    \mathcal{P}_k = \{ p_{k,1}, \dots, p_{k,m_k} \}, \qquad p_{k,j} = r(x_j^k).
    \label{eq:pixel_bank}
\end{equation}
Nearest clients are then selected based on the Euclidean distance in pixel space. While this approach is model-agnostic, it is sensitive to nuisance factors such as contrast or illumination, computationally expensive, and empirically less accurate in assigning clients relevant to a test sample.
Across all datasets, feature-space similarity significantly outperforms raw-pixel assignment in top-$k$ client assignment accuracy and computational efficiency. The results are provided in the supplementary material. By integrating representation-aware client assignment, TrustFed ensures that each test sample draws calibration information from the most relevant clients, which is crucial for producing \emph{efficient, accurate, and privacy-preserving} conformal prediction sets in heterogeneous federated learning settings.

\subsection*{Soft-nearest threshold aggregation}
While representation-aware client assignment identifies a localized neighborhood of relevant clients for a given test sample, it remains nontrivial to aggregate their calibration information in a manner that is both robust to assignment uncertainty and efficient in prediction set size. To this end, we propose a \emph{soft-nearest threshold aggregation} strategy for determining the test-sample-specific conformal threshold.
For a test input $x_{\text{test}}$, let $\mathcal{N}_k(x_{\text{test}})$ denote the set of the $\mathcal{N}_k$ nearest clients in the learned feature space, as determined by the representation-aware client assignment described in the previous subsection. Each client $j \in \mathcal{N}_k$ independently computes a local conformal threshold $\tau_j$ using its private calibration set, without sharing calibration scores or labels.
Rather than selecting a single client or pooling calibration scores across clients, we aggregate these local thresholds using a conservative max-operator:
\begin{equation}
\tau(x_{\text{test}}) = \max_{j \in \mathcal{N}_k(x_{\text{test}})} \tau_j ,
\label{eq:max_tau}
\end{equation}
which yields a test-dependent threshold that adapts to the local data heterogeneity. We refer to this procedure as \emph{soft-nearest threshold aggregation}. The resulting conformal prediction set is then defined as
\begin{equation}
\Gamma(x_{\text{test}}) = \{ y \in \{1, \dots, C\} : s(x_{\text{test}}, y) \leq \tau(x_{\text{test}}) \},
\label{eq:prediction_set}
\end{equation}
where $s(x,y)$ denotes the nonconformity score.

This aggregation strategy is motivated by the observation that client assignment based on learned representations is inherently noisy under real-world heterogeneity. Selecting a single nearest client ($\mathcal{N}_k = 1$) can lead to overly optimistic thresholds and potential under-coverage if the assignment is incorrect due to overlapping client distributions or representation noise. By incorporating multiple nearby clients, soft-nearest aggregation mitigates this risk by safeguarding coverage against misassignment.
The choice of $\mathcal{N}_k$ governs a fundamental coverage-efficiency trade-off. Smaller values of $\mathcal{N}_k$ tend to produce compact prediction sets but are more sensitive to assignment errors. Increasing $\mathcal{N}_k$ improves robustness by accounting for uncertainty in client relevance, at the cost of potentially larger thresholds and hence less efficient prediction sets. In the limit, very large $\mathcal{N}_k$ recovers a conservative global behavior.
{This smoothing mechanism introduces a mild form of data-dependent conditioning, since the effective calibration neighborhood depends on the test representation, departing from the classical split conformal setting where the calibration set is fixed a priori. While such adaptive selection could, in principle, induce selection bias, TrustFed mitigates this risk by taking the maximum threshold across the selected neighborhood; the effective threshold is guaranteed to be no smaller than that of any individual neighboring client, thereby guarding against undercoverage even under imperfect client assignment.}
In practice, we select $\mathcal{N}_k$ empirically by choosing the smallest value that achieves the target coverage level $1-\alpha$ with a small safety margin. Importantly, this aggregation mechanism preserves privacy and communication efficiency: only scalar thresholds are exchanged, and no calibration scores, embeddings, or labels are revealed.
By explicitly accounting for uncertainty in client assignment, soft-nearest threshold aggregation constitutes a key methodological contribution of TrustFed. Together with representation-aware client assignment, it enables adaptive, uncertainty-aware conformal prediction in heterogeneous federated learning settings without violating privacy constraints.
\subsection*{Network architecture}
While the proposed approach can be used with any network architecture of choice, we have used CortiNet \cite{kumar2026cortinetphysicsperceptionhybridcorticalinspired} in this study. It is a physics-perception hybrid, wavelet-based deep learning model that explicitly exploits the multi-resolution and frequency-localized structure of medical images. Unlike conventional convolutional neural networks that implicitly learn the features through stacked convolution and pooling operations, CortiNet integrates the discrete wavelet transform as a core architectural component, thereby embedding domain-informed inductive bias directly into the learning pipeline. The discrete wavelet transform enables decomposition of input images into complementary low and high-frequency sub-bands. The input image is decomposed into an approximation (low-frequency) component that captures global anatomical structure and detail (high-frequency) components that encode details such as edges and textures. At each wavelet decomposition stage, the input feature maps are decomposed into approximation and detail components, allowing the model to jointly capture global anatomical context and fine-grained textural variations. This explicit frequency separation is well-suited for medical imaging, where the diagnostically relevant structures coexist with noise-dominated high-frequency artifacts. 
Following wavelet decomposition, CortiNet employs a dual-stream encoding strategy inspired by parallel processing pathways in the human visual cortex. The low-frequency approximation component is processed by the structural stream, which focuses on learning coarse-scale morphology and anatomical context. In parallel, the high-frequency detail components are processed by a perceptual stream, which learns texture and edge information.
The sub-band representations are processed through lightweight convolutional blocks. The representations learned by the structural and perceptual streams are integrated via a late-fusion mechanism. This late-stage integration preserves the semantic distinction between structure and detail components while enabling joint reasoning over complementary clues. Moreover, this design prevents high-frequency noise from contaminating global structural representations at early stages of processing. A key feature of CortiNet is its noise-aware adaptive inference mechanism, which is particularly relevant in real-world settings. Since high-frequency components are more susceptible to noise, CortiNet uses a calibration-based criterion to assess the reliability of the structural stream. When structural predictions are deemed reliable, inference is performed using the structural pathway alone. Otherwise, the perceptual stream is activated and fused with structural features. This adaptive strategy favours stable low-frequency reasoning under noisy conditions while selectively incorporating high-frequency detail when beneficial. 

This architectural design results in a compact, parameter-efficient architecture. By operating on structured frequency representations rather than raw pixels and by using lightweight convolutional encoders, CortiNet achieves competitive performance with a significantly reduced parameter footprint and low inference latency. Further, the explicit frequency decomposition increases robustness to common imaging noise and artifacts, which are prevalent in the medical domain. These advantages make CortiNet a suitable choice for a decentralized setting implemented in this work.

\subsection*{Algorithmic Overview of FedTrust}
We now summarize the complete FedTrust pipeline, which integrates model training, calibration, and test-time uncertainty quantification into a single end-to-end framework. Overall, the proposed method combines three key components: 
(i) privacy-preserving federated training of a shared global model, 
(ii) per-client conformal calibration coupled with local feature-bank construction, and 
(iii) test-time, sample-adaptive prediction-set construction via representation-aware neighbour selection and soft-nearest threshold aggregation.
The procedure proceeds in three stages. First, a global model is trained using standard federated optimization, where each client performs local updates on its private training data and only model parameters are shared with the central server. This stage yields a global representation function that is consistent across clients while respecting data privacy. Second, each client performs local conformal calibration using a held-out calibration split. During this step, nonconformity scores and client-specific conformal thresholds are computed locally, and a feature bank is constructed by storing learned representations of the calibration samples. {The representation function $\phi(\cdot)$ is fully fixed after federated training and before calibration begins. The calibration scores, feature banks, and conformal thresholds are all computed using the same frozen global model. This ensures consistency between the feature structure used during calibration and the feature structure encountered at test time.}
Importantly, neither calibration data nor nonconformity scores are communicated to the server, ensuring privacy preservation. 
Finally, at inference time, the framework constructs prediction sets in a sample-adaptive manner. For a given test input, its feature representation is broadcast to the clients, which return only scalar distance values measuring proximity to their local feature banks. Based on these distances, the server identifies a small neighbourhood of the most relevant clients and aggregates their local conformal thresholds using a soft-nearest strategy. The resulting adaptive threshold is then used to generate a conformal prediction set that satisfies the desired coverage guarantee while remaining compact under heterogeneous data distributions.

Algorithm~\ref{algo:algorithm} details the complete FedTrust procedure.

\begin{algorithm}[H]
\SetKwInput{Input}{Input}
\SetKwInput{Output}{Output}
\SetAlgoLined
\LinesNumbered
\Input{Global model $f_\theta$ with parameters $\theta$; $R$ federated training rounds; $K$ clients; local epochs $E$;
local training sets $\{(\hat x^k_i,\hat y^k_i)\}_{i=1}^{m_k}$ and calibration sets $\{(x^k_i,y^k_i)\}_{i=1}^{n_k}$;
conformal score function $S:\Delta^J\!\to\!\mathbb{R}_+$; error level $\alpha$, and hyperparameters}
\textit{// Step 1: Federated training}\\
Train the global model via federated optimization: \\
$f \leftarrow \texttt{FedOpt}\bigl(f_\theta,\ \{(\hat x^k_i,\hat y^k_i)\}_{i,k},\ R, E\bigr)$ \hfill $\triangleright$ Eq.~\eqref{eq:global_objective}\\

\textit{// Step 2: Local calibration and feature-bank construction}\\
\For{$k \in \{1,2,\dots,K\}$}{
    \For{$i \in \{1,2,\dots,m_k\}$}{
        Compute nonconformity scores on local calibration data:\\
        $s_i^k \gets S\bigl(f(x_i^k),y_i^k\bigr)$ \hfill $\triangleright$ Eq.~\eqref{eq:score}
    }
    Compute client-specific conformal threshold:\\
    $\hat\tau^{(k)} = S_{(\lceil (1-\alpha)(m_k+1)\rceil)}$ \hfill $\triangleright$ Eq.~\eqref{eq:empirical_quantile}\\
    Construct local feature bank:\\
    $\mathcal{B}_k = \{\,b_{k,1},\dots,b_{k,m_k}\} \subset \mathbb{R}^d$ \hfill $\triangleright$ Eq.~\eqref{eq:feature_bank}
}
\textit{// Step 3: Sample-adaptive prediction set construction}\\
\For {$i \in \{1,2,\dots,n_t\}$}{
    Encode test input to obtain feature representation:
    $f_i \gets \phi(x_i)$\\
    \For{$k \in \{1,\dots,K\}$}{
        Compute nearest-feature distance:\\
        $d_{i,k} = \min_{1 \le j \le m_k} \| f_i - b_{k,j} \|_2$ \hfill $\triangleright$ Eq.~\eqref{eq:feature_min_dist}
    }
    Identify $k$ nearest clients in feature space:\\
    $\mathcal{N}_k \gets \argsort(d_{i,1},\dots,d_{i,K})[1:k]$\\
    Aggregate thresholds via soft-nearest rule:
    $\tau \gets \max_{j \in \mathcal{N}_k} \tau_j$ \hfill $\triangleright$ Eq.~\eqref{eq:max_tau}\\
    Construct conformal prediction set:\\
    $\Gamma_\alpha(x_i) = \{\, y \in \{1, \ldots, C\} : S(x_i,y_i) \le \tau \,\}$ \hfill $\triangleright$ Eq.~\eqref{eq:prediction_set}
}

\Output{Set-valued predictor $\Gamma_\alpha(\cdot)$}
\caption{FedTrust (Sample-Adaptive Federated Conformal Prediction)}
\label{algo:algorithm}
\end{algorithm}

\section{Results}
\label{sec:results}
We evaluated the proposed TrustFed framework on a large and heterogeneous collection of medical imaging datasets designed to reflect the scale, diversity, and imbalance encountered in real-world, multi-institutional healthcare systems. 
\begin{figure}[!ht]
    \centering
    \includegraphics[width=\textwidth]{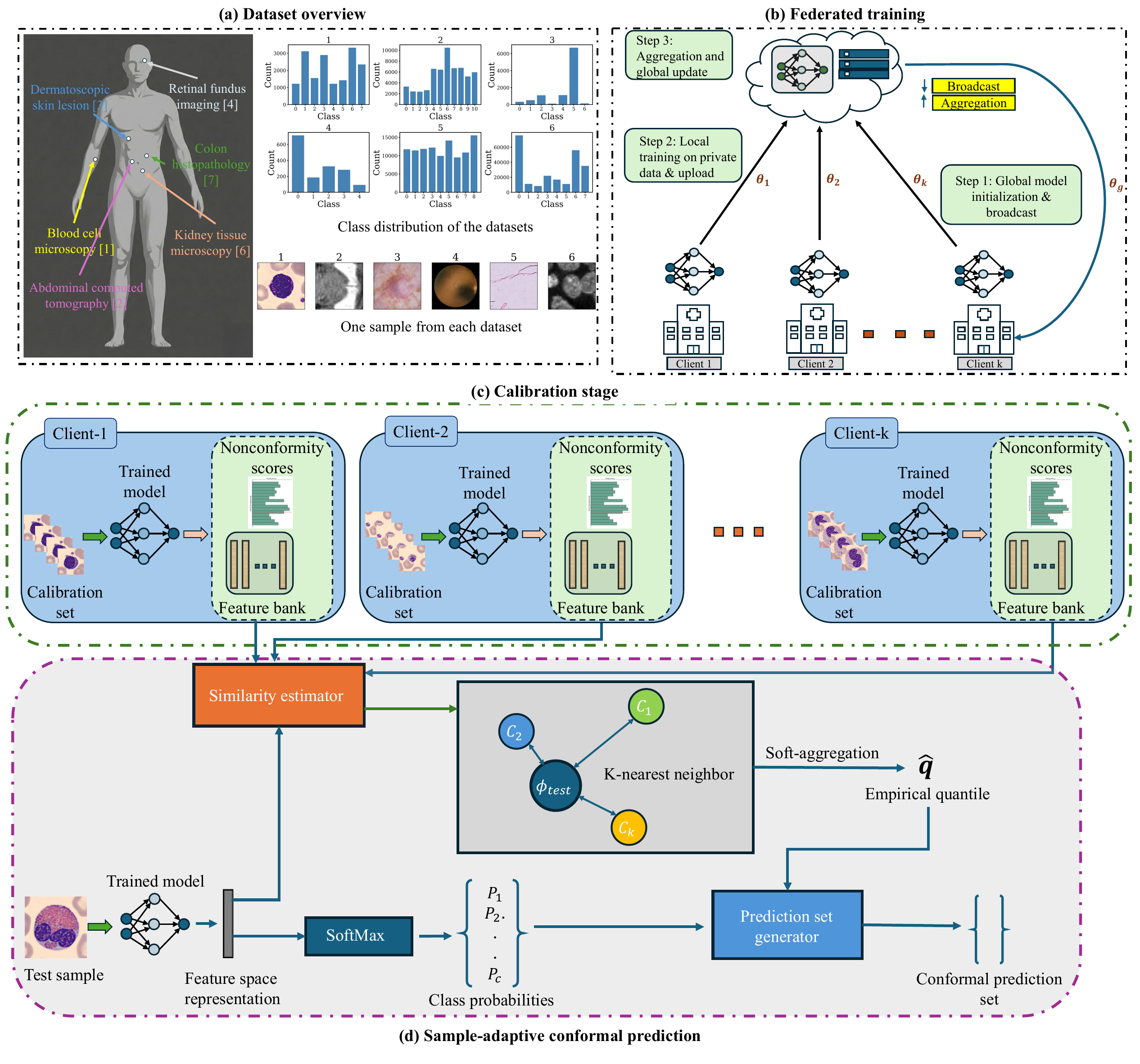}
    \caption {\textbf{Overview of sample-adaptive federated conformal prediction framework.} \textbf{a}	Dataset overview: Six dataset involving blood cell microscopy, abdominal computed tomography, dermatoscopic skin lesion, Retinal fundus imaging, kidney tissue microscopy, and colon histopathology are considered to cover the diverse medical imaging modalities. The bar plots show the class distributions of each dataset, while example images illustrate one representative sample per dataset, highlighting the visual characteristics. A schematic human body is used to symbolically represent each dataset. \textbf{b}	Federated training: A global model is initialized and broadcast to all clients. Each client performs local training on its private data and uploads the updated parameters, which are then aggregated to update the global model. \textbf{c}	Calibration stage: Each client uses a held-out calibration set to compute nonconformity scores and stores them locally. \textbf{d}	Sample-adaptive conformal prediction: For a given test sample, its feature space representation is first obtained via the trained model and passed through a similarity estimator. Using K-nearest neighbor search across client feature banks, the most relevant calibration instances are selected. The threshold empirical quantile is obtained by soft nearest threshold aggregation strategy. This quantile, together with the model’s class probabilities from softmax, is used by the prediction set generator to construct a conformal prediction set.}
    \label{fig:figure1}
\end{figure}
As summarized in \Cref{fig:figure1}a, our study considers six clinically distinct imaging tasks spanning blood cell microscopy, retinal fundus imaging, abdominal computed tomography, dermatoscopic skin lesion analysis, colon histopathology, and kidney tissue microscopy \cite{yang2023medmnist}. Collectively, these datasets comprise over 430,000 images, with individual dataset sizes ranging from approximately 1,600 to 236,000 samples and diagnostic class cardinalities spanning 5 to 11 classes, capturing a wide spectrum of visual complexity and clinically relevant class imbalance patterns. The overall federated training and uncertainty estimation workflow is illustrated in \Cref{fig:figure1}b–d, where a global model is trained through privacy-preserving client updates (\Cref{fig:figure1}b), client-specific calibration statistics are computed locally using held-out data (\Cref{fig:figure1}c), and sample-adaptive conformal prediction is performed at inference by selecting representation-similar calibration instances and aggregating thresholds via a soft-nearest strategy to construct prediction sets (\Cref{fig:figure1}d). 
For all experiments, data were distributed across multiple federated clients to emulate institutional data silos with non-identically distributed class distributions, and no raw images or labels were exchanged at any stage. TrustFed was evaluated against existing federated and centralized uncertainty quantification baselines using empirical coverage relative to nominal guarantees and the average size of conformal prediction sets, enabling a rigorous assessment of statistical validity and predictive efficiency across diverse medical imaging tasks.

\subsection{TrustFed achieves nominal coverage across all tasks}
We first assess whether TrustFed satisfies its primary objective of delivering statistically valid uncertainty estimates in heterogeneous federated settings. Across all six medical imaging tasks and federated configurations, TrustFed consistently achieves empirical coverage closely matching the prescribed nominal levels, demonstrating reliable uncertainty calibration under non-identically distributed data and severe class imbalance (\Cref{fig:figure3,fig:figure4}). This behavior is observed uniformly across imaging modalities, dataset scales, and class cardinalities, including tasks characterized by rare diagnostic categories, where existing federated uncertainty quantification approaches exhibit systematic under-coverage. In contrast, baseline methods show deviations from nominal coverage, particularly as data heterogeneity and imbalance increase, highlighting their sensitivity to client-specific distributional shifts. Notably, the required coverage is achieved without inflating prediction sets, indicating that coverage is not obtained through overly conservative uncertainty estimates. These results confirm that integrating representation-aware client similarity and sample-adaptive threshold aggregation enables TrustFed to preserve the theoretical guarantees of conformal prediction in practical federated healthcare scenarios. Collectively, this finding establishes that statistically valid uncertainty quantification can be achieved at scale in privacy-preserving, multi-institutional medical imaging, reinforcing the necessity of principled uncertainty guarantees for trustworthy clinical deployment.

\subsection{Impact of representation-aware client assignment}
We next investigate the contribution of representation-aware client assignment to the uncertainty calibration performance of TrustFed. In heterogeneous federated settings, client populations often differ substantially in both data distribution and feature representation, rendering uniform or random client aggregation ineffective for conformal calibration. By leveraging model-internal feature representations to group clients with similar data characteristics (\Cref{fig:figure2}a), TrustFed enables calibration to be performed within more homogeneous subsets of the federated population, while preserving data privacy and communication efficiency. Across all medical imaging tasks, representation-aware client assignment leads to systematic improvements in both coverage stability and prediction set efficiency compared to client-agnostic calibration strategies. In particular, we observe a marked reduction in coverage variability across clients and classes, with the largest gains arising in datasets exhibiting pronounced distributional skew and rare diagnostic categories. These findings indicate that aligning calibration with representation-level similarity mitigates the adverse effects of client heterogeneity, allowing TrustFed to better approximate the exchangeability assumptions underlying conformal prediction in practical federated healthcare scenarios. {Importantly, this improvement can be understood through the lens of exchangeability, which underpins conformal prediction. Rather than assuming global exchangeability across all clients, TrustFed adopts a weaker and more realistic assumption of approximate exchangeability within representation neighborhoods. Calibration and test samples that are close in the learned feature space are treated as approximately exchangeable, even when originating from different institutions. This localized notion of exchangeability is particularly appropriate in federated healthcare settings, where institutions with similar patient populations and imaging protocols tend to generate statistically compatible data distributions.} The benefits of representation-aware assignment persist across dataset scales and imaging modalities, demonstrating that its effectiveness is not limited to a specific task or cohort size. Collectively, these results highlight representation-aware client assignment as a key enabler of statistically reliable and efficient uncertainty quantification in federated learning, providing a principled mechanism to reconcile privacy preservation with the need for robust calibration in multi-institutional medical imaging.

\subsection{Soft-nearest threshold aggregation mitigates assignment uncertainty}
While representation-aware client assignment improves calibration by grouping clients with similar model representations, assignment errors can occur due to imperfect similarity estimation or limited client data. To address this, TrustFed incorporates a soft-nearest threshold aggregation strategy, which smooths the influence of calibration instances across nearby clients rather than relying solely on exact nearest neighbors. 
This mechanism ensures that prediction sets remain statistically valid even when client assignments are noisy. Across all six medical imaging tasks, we observe that soft-nearest aggregation substantially reduces the variability in empirical coverage that arises from imperfect client grouping. In particular, tasks with high inter-client heterogeneity or rare diagnostic categories show pronounced improvements, with TrustFed maintaining coverage closer to nominal levels than both global aggregation and strict nearest-neighbor approaches. Importantly, this enhanced robustness is achieved without increasing the average size of prediction sets (\Cref{fig:figure3}c), demonstrating that soft-nearest aggregation mitigates assignment uncertainty while preserving practical efficiency. These results indicate that incorporating a soft aggregation mechanism substantially improves robustness for deployment in heterogeneous clinical environments, where data heterogeneity and client-specific biases are unavoidable. By stabilizing coverage under imperfect client assignments, TrustFed provides more reliable and trustworthy uncertainty estimates for high-stakes medical decision-making.

\subsection{Optimal no. of neighbors improves accuracy}
\begin{figure}[!ht]
    \centering
    \includegraphics[width=\textwidth]{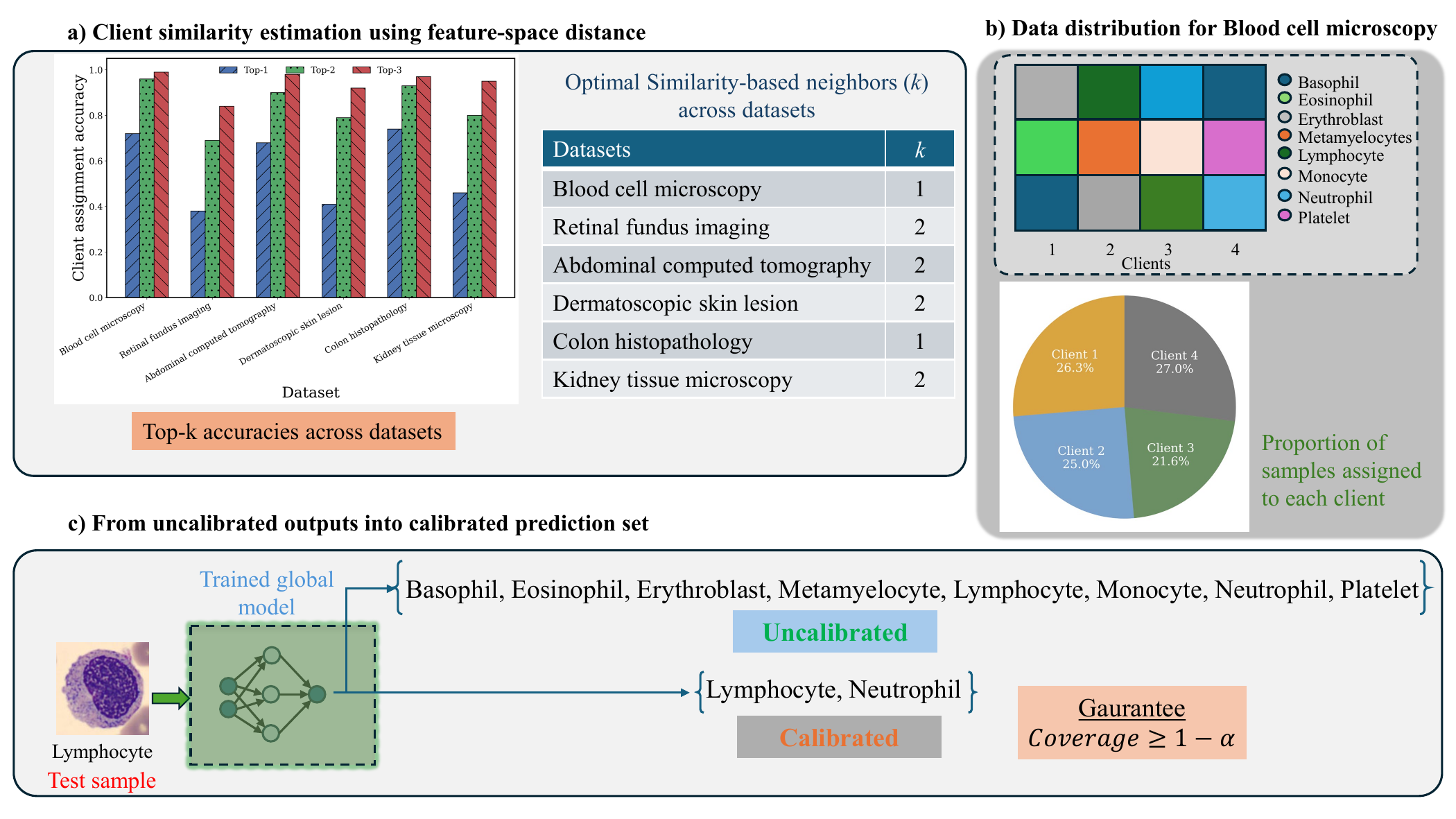}
    \caption{\textbf{Client similarity assignment and calibration of prediction set under class imbalance.} \textbf{a} Client assignment accuracy is computed using Euclidean distance in the feature space. Accuracy improves with larger top-k neighbor sets, ensuring reliable client similarity estimation. \textbf{b} The class distribution across clients and the proportion of samples per client are illustrated for blood cell microscopy. \textbf{c} A test sample (lymphocyte) passed through the trained global model initially yields an uncalibrated set of possible labels. Conformal prediction narrows this to a calibrated prediction set with guaranteed coverage $\geq 1-\alpha$}
    \label{fig:figure2}
\end{figure}

The choice of the number of neighbors $N_k$ in the sample-adaptive calibration stage plays a critical role in balancing statistical validity and predictive efficiency. When $N_k=1$, the nonconformity score for a test sample is determined solely by its nearest client in representation space. While this highly localized calibration can yield very compact prediction sets, it is also sensitive to representation noise and overlapping client distributions, leading to potential under-coverage when nearest-client assignments are imperfect. This behavior is particularly pronounced in heterogeneous federated settings, where client boundaries are often diffuse rather than well separated. Increasing $N_k$ improves robustness by incorporating calibration information from a broader neighborhood of representation-similar clients. As $N_k$ grows, the resulting threshold becomes less sensitive to individual assignment errors, thereby stabilizing empirical coverage across clients and classes. However, excessively large values of $N_k$ lead to increasingly conservative thresholds, which inflate prediction set sizes and reduce their clinical interpretability (\Cref{fig:figure3}c). This reflects a fundamental coverage–efficiency trade-off, where larger neighborhoods guarantee safety at the expense of informativeness. Across all medical imaging tasks and federated configurations, we observe the existence of an intermediate optimal range of $N_k$ that achieves nominal coverage while maintaining compact prediction sets (\Cref{fig:figure2}a). This regime consistently outperforms both the extreme local ($N_k=1$) and global (large $N_k$) calibration strategies. These findings demonstrate that adaptive neighborhood selection is essential for effective federated conformal prediction, and that principled tuning of $N_k$ enables TrustFed to reconcile robustness and efficiency under realistic clinical heterogeneity.

\subsection{TrustFed produces compact prediction sets under heterogeneity}
We next investigate the contribution of representation-aware client assignment to the uncertainty calibration performance of TrustFed. In heterogeneous federated settings, client populations often differ substantially in both data distribution and feature representation, rendering uniform or random client aggregation ineffective for conformal calibration. By leveraging model-internal feature representations to group clients with similar data characteristics, TrustFed enables calibration to be performed within more homogeneous subsets of the federated population, while preserving data privacy and communication efficiency. Across all medical imaging tasks, representation-aware client assignment leads to systematic improvements in both coverage stability and prediction set efficiency compared to client-agnostic calibration (\Cref{fig:figure3}a, and \Cref{fig:figure4} a). In particular, we observe a marked reduction in coverage variability {across clients and classes}, with the largest gains arising in datasets exhibiting pronounced distributional skew and rare diagnostic categories. These findings indicate that aligning calibration with representation-level similarity mitigates the adverse effects of client heterogeneity, allowing TrustFed to better approximate the exchangeability assumptions underlying conformal prediction in practical federated healthcare scenarios. Notably, the benefits of representation-aware assignment persist across dataset scales and imaging modalities, demonstrating that its effectiveness is not limited to a specific task or cohort size. Collectively, these results highlight representation-aware client assignment as a key enabler of statistically reliable and efficient uncertainty quantification in federated learning, providing a principled mechanism to reconcile privacy preservation with the need for robust calibration in multi-institutional medical imaging.

\subsection{Performance under extreme class imbalance}
\begin{figure}[!ht]
    \centering
    \includegraphics[width=\textwidth]{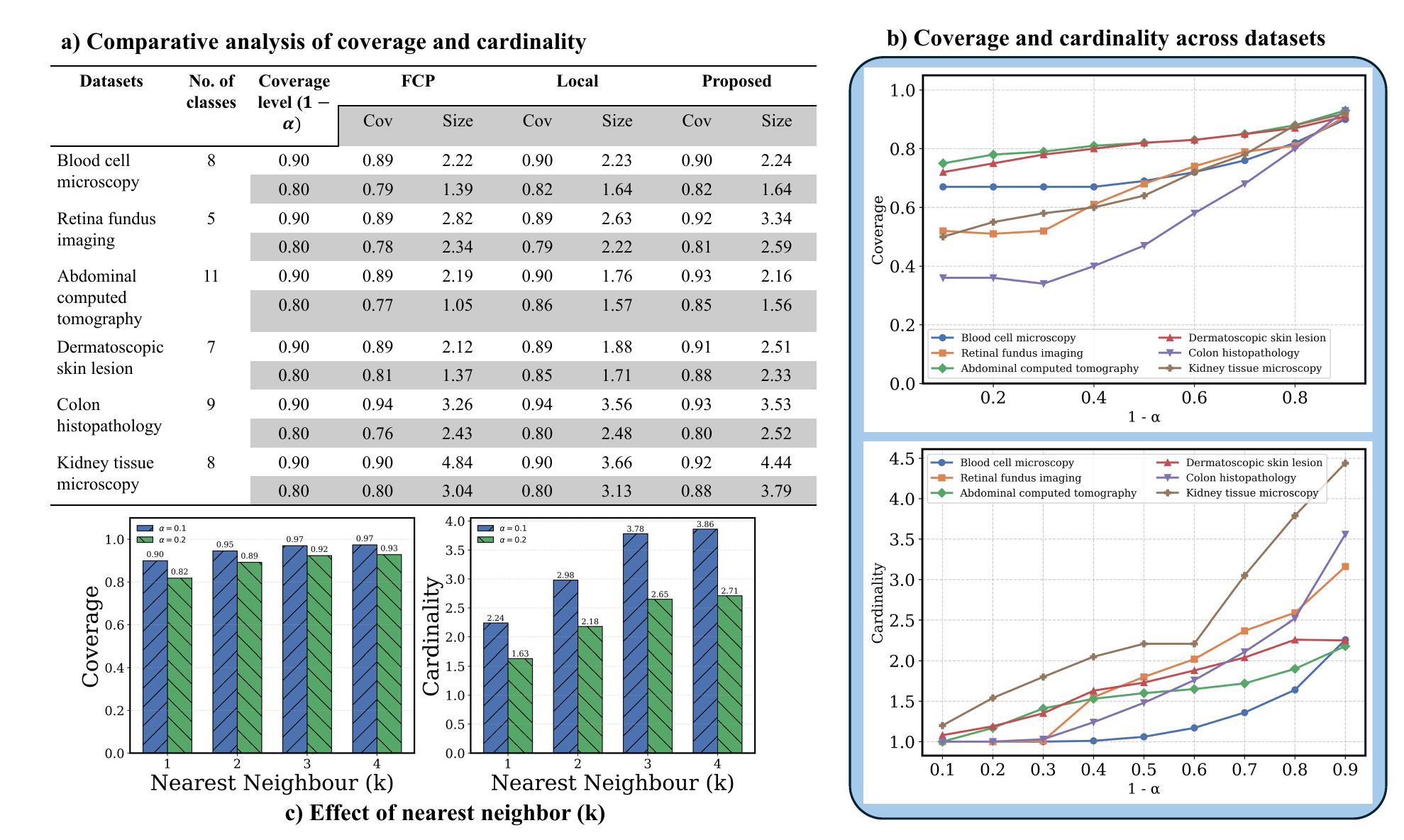}
    \caption{\textbf{Comparative coverage and cardinality analysis across the six datasets under class imbalance.} \textbf{a} The table compares FCP, Local, and the Proposed approach across six datasets discussed before under coverage levels $(1-\alpha)=0.9$ and 0.8. It shows that the proposed method consistently achieves empirical coverage close to the nominal level while maintaining competitive prediction set cardinality across datasets. \textbf{b} This plot shows coverage and cardinality as function of $1-\alpha$. \textbf{c} This bar chart shows coverage and cardinality variation for $\alpha=0.9$ and 0.8 under different nearest neighbor (k) values. Larger k improves coverage at the cost of slightly increased cardinality, indicating a trade-off between coverage and set size.}
    \label{fig:figure3}
\end{figure}

Clinical datasets often contain rare diagnostic categories, where accurate and reliable uncertainty estimates are particularly critical. To evaluate TrustFed under these challenging conditions, we analyzed its performance on tasks exhibiting severe class imbalance, with minority classes representing only a small fraction of the overall dataset. Across all six medical imaging tasks, TrustFed consistently maintains empirical coverage near nominal levels for both majority and minority classes, demonstrating robustness to extreme distributional skew that commonly undermines conventional federated and centralized uncertainty quantification methods (\Cref{fig:figure3}a). Importantly, the average size of the prediction sets for rare classes remains comparable to or smaller than that for majority classes, highlighting that TrustFed achieves statistically valid coverage without resorting to overly conservative predictions (\Cref{fig:figure3}b). In contrast, baseline approaches often produce excessively large prediction sets for underrepresented classes or fail to meet coverage guarantees, limiting their clinical utility. These findings underscore that TrustFed’s combination of representation-aware client assignment and soft-nearest threshold aggregation enables principled uncertainty estimation even in the most challenging, high-stakes clinical scenarios. Consequently, TrustFed provides reliable, interpretable prediction sets for rare diagnostic categories, a key requirement for safe deployment in multi-institutional medical imaging workflows.

\subsection{Performance under sample imbalance}
\begin{figure}[!ht]
    \centering
    \includegraphics[width=\textwidth]{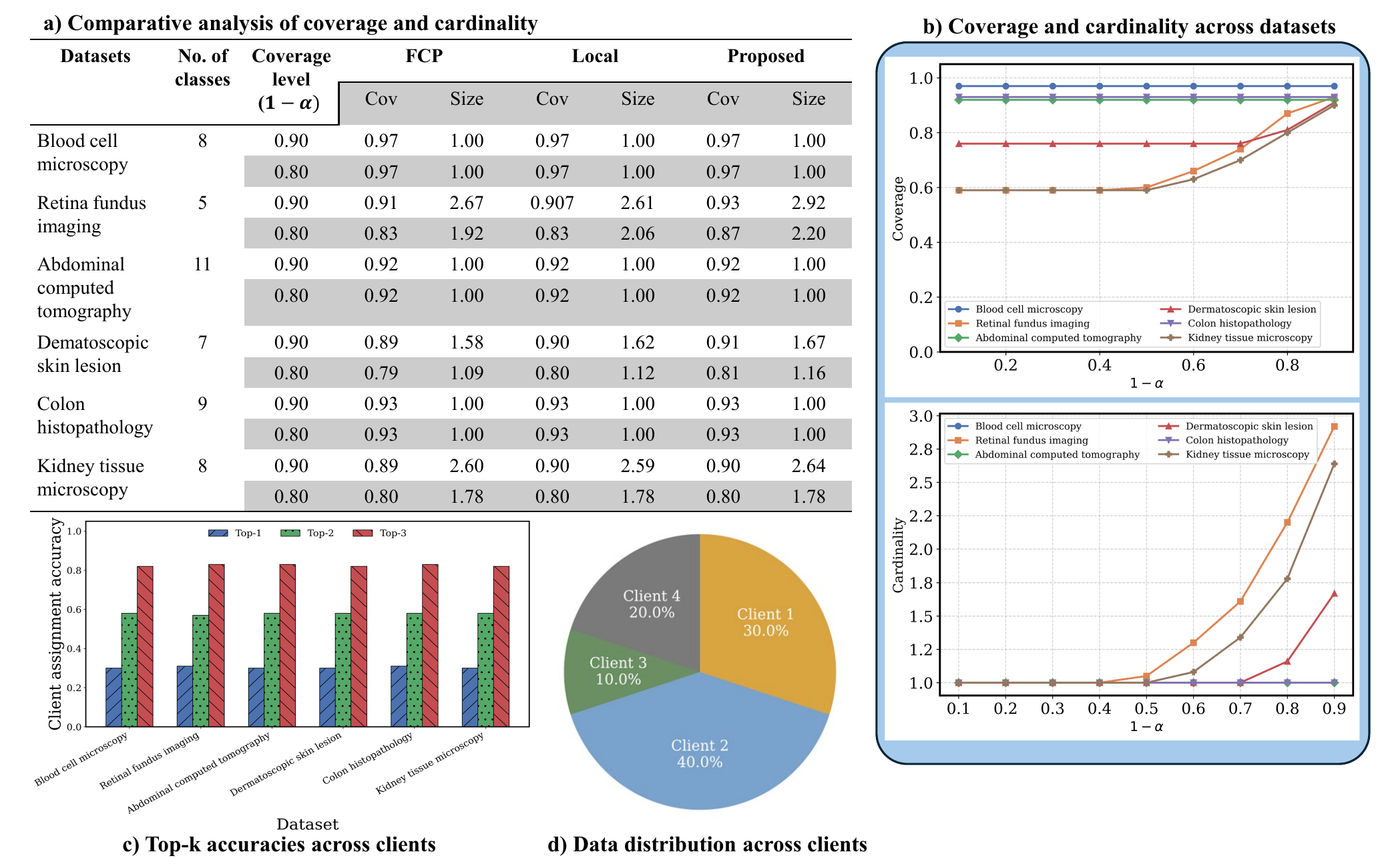}
    \caption{\textbf{Comparative coverage and cardinality analysis across the datasets under sample imbalance.} \textbf{a} The table compares the coverage and cardinality across all six datasets at specified confidence levels $(1-\alpha)=0.9$ and 0.8. \textbf{b} This plot shows that the FedTrust achieves the required coverage for various values of $\alpha$. \textbf{c} This bar chart shows top-k client assignment accuracy across datasets. \textbf{d} This pie chart shows the distribution of samples across clients, while maintaining the same class distribution for each client.
    } 
    \label{fig:figure4}
\end{figure}

In real-world scenarios, we frequently encounter sample imbalance across participating clients, with some clients having substantially fewer training samples than others. To assess the robustness of the FedTrust under these conditions, we evaluated its performance under skewed sample distribution across the clients as shown in \Cref{fig:figure4}. TrustFed consistently attains the required coverage corresponding to various values of confidence level across all datasets, despite substantial variation in the number of samples present at each client (\Cref{fig:figure4}a,d). 
% \textcolor{blue}{We note that clients with very small calibration splits produce noisier quantile estimates. However, two factors mitigate this effect in practice. First, the representation-aware client assignment favors clients whose feature distributions are similar to the test sample, and clients with very few samples tend to have sparser and less representative feature banks, making them less likely to be selected as nearest neighbors. Second, the max-aggregation biases errors toward over-coverage rather than under-coverage, which is the safer direction for clinical deployment.}
{As shown in \Cref{fig:figure4}c, TrustFed's representation-aware client assignment achieves high top-1 accuracy across most datasets, with top-2 and top-3 accuracies approaching near-perfect levels, confirming that test samples are reliably routed to the most relevant clients.}
Further, we examined the size of the resulting prediction sets to assess whether robustness under sample imbalance is achieved through overly conservative prediction set. As shown in \Cref{fig:figure4}a, the valid coverage is achieved without inflating the prediction sets. The coverage and cardinality for various confidence levels are obtained to assess the robustness of TrustFed to increasing confidence requirements in the presence of sample imbalance (\Cref{fig:figure4}b). It shows that the proposed approach achieves the required coverage. 

\subsection*{Discussions}
\label{sec:discussions}
In this study, we introduce Federated Conformal Prediction (TrustFed), a privacy-preserving framework that enables statistically valid uncertainty quantification in federated medical imaging at scale, addressing a fundamental limitation that has hindered the reliable deployment of federated learning in healthcare. By validating TrustFed across over 430,000 medical images spanning six clinically distinct imaging modalities, our work represents one of the most comprehensive investigations of uncertainty-aware federated learning in medical imaging to date. The federated paradigm enables this breadth of evaluation by allowing training and calibration across institutionally siloed datasets without centralizing sensitive patient data. Unlike most prior federated learning studies that primarily focus on predictive accuracy, our work directly addresses the trustworthiness and reliability of model outputs, which are essential prerequisites for clinical decision-making. By demonstrating that rigorous, distribution-free coverage guarantees can be preserved under realistic heterogeneity, severe class imbalance, and client-specific biases, this study provides evidence that uncertainty-aware federated learning can be practically viable and clinically meaningful paradigm, with the potential to support safe and generalizable deployment of machine learning systems across healthcare institutions.
Our results demonstrate that access to large and diverse data alone is insufficient to ensure reliable uncertainty estimation in federated healthcare systems. While federated learning enables collaboration across institutions, naïvely aggregating calibration information across heterogeneous clients leads to either systematic under-coverage or overly conservative prediction sets, particularly in the presence of rare diagnostic categories. This observation parallels findings in the broader machine learning literature, where performance gains from increased data volume can plateau or even degrade when heterogeneity, label noise, and site-specific biases are not explicitly addressed. In our experiments, TrustFed consistently achieves nominal coverage and compact prediction sets not because of increased data alone, but due to selective, representation-aware utilization of calibration information. This highlights a key insight: in federated settings, how calibration data are leveraged is as important as how much data are available. Our findings underscore that principled uncertainty quantification in healthcare requires mechanisms that explicitly account for inter-client variability, rather than assuming that scale alone will yield trustworthy predictions.
A central contribution of this work lies in addressing a long-standing and largely unresolved challenge in federated learning: how to achieve statistically valid uncertainty quantification in the presence of client heterogeneity and class imbalance, without compromising privacy. While conformal prediction offers rigorous, distribution-free coverage guarantees in centralized settings, its direct application to federated environments is fundamentally limited by heterogeneous client distributions and misaligned calibration data. The proposed representation-aware client assignment directly addresses this gap by enabling calibration to be conditioned on model-internal representations rather than client identity or global statistics, thereby aligning uncertainty estimation with the underlying data-generating mechanisms across institutions. Complementing this, the soft-nearest threshold aggregation strategy mitigates the inevitable uncertainty arising from imperfect client similarity estimation by smoothing threshold selection across neighboring clients, ensuring robustness without inflating prediction sets. 

To the best of our knowledge, this is the first study to integrate representation-aware similarity and soft aggregation into a federated conformal prediction framework, transforming conformal prediction from a theoretically appealing but fragile tool into a practically deployable mechanism for trustworthy uncertainty estimation in federated healthcare AI.
Beyond the specific methodological advances introduced in this work, our findings have broader implications for the design and deployment of trustworthy artificial intelligence in privacy-sensitive domains. By demonstrating that statistically valid and efficient uncertainty quantification can be achieved in federated settings under severe heterogeneity and class imbalance, this study challenges the prevailing assumption that strong uncertainty guarantees necessarily require centralized data access or overly conservative prediction strategies. The proposed representation-aware client assignment and soft-nearest threshold aggregation establishes a general strategy for adapting classical statistical guarantees to modern, distributed learning paradigms. While our evaluation focuses on medical imaging, the principles underlying TrustFed are broadly applicable to other high-stakes domains where data decentralization, distribution shift, and reliability are fundamental concerns, including personalized medicine, distributed sensing, finance, and federated decision support systems. More broadly, our results suggest that privacy preservation, statistical rigor, and practical usability need not be competing objectives, but can instead be jointly addressed through principled algorithmic design. We anticipate that this work will motivate future efforts at the intersection of federated learning, uncertainty quantification, and trustworthy AI, accelerating the transition of federated models from experimental deployments to real-world decision-making systems.
Although our study is motivated by challenges in multi-institutional healthcare, the principles underlying TrustFed extend well beyond medical imaging and clinical decision-making. Many real-world federated learning deployments, spanning domains such as autonomous systems, smart infrastructure monitoring, climate and environmental modeling, finance, and personalized recommender systems, operate under similarly severe conditions of data heterogeneity, non-identically distributed client data, and class imbalance, where reliable uncertainty quantification is essential for risk-aware decision-making. Moreover, the differential privacy constraints impose additional challenges for calibration and uncertainty estimation \cite{geyer2017differentially}.
The representation-aware client assignment and soft-nearest threshold aggregation strategies introduced in TrustFed are model-agnostic and task-agnostic, relying solely on learned feature representations rather than domain-specific assumptions. As such, they can be readily integrated with a wide range of architectures, data modalities, and learning objectives commonly used in federated settings outside healthcare. More broadly, by demonstrating that statistically guaranteed uncertainty can be preserved under realistic federated constraints, our work suggests that conformal prediction can serve as a general-purpose statistical layer for trustworthy federated intelligence. This suggests that uncertainty-aware federated learning can serve as a statistical layer for safety-critical and decision-sensitive applications across scientific, industrial, and societal domains, where privacy, heterogeneity, and reliability must be addressed simultaneously.
This study has several limitations that point to important directions for future research. First, while TrustFed is evaluated across multiple medical imaging tasks, each task involves a single imaging modality, and extending the framework to multi-modal federated settings, where heterogeneous data types are distributed across institutions, remains an important open challenge. Second, the current formulation of TrustFed is focused on classification tasks, and extending conformal uncertainty guarantees to regression, structured prediction, and segmentation outputs within federated environments represents a promising avenue for broadening its applicability. Third, the choice of neighborhood size in the soft-nearest threshold aggregation introduces an explicit trade-off between robustness and efficiency, which is currently selected empirically; developing adaptive or theoretically guided strategies for neighborhood selection could further improve both coverage stability and prediction set compactness. Addressing these challenges will strengthen the generality and applicability of uncertainty-aware federated learning and enable deployment in increasingly complex, real-world decision-making scenarios.
In summary, this work advances federated learning beyond privacy preservation alone by establishing statistically guaranteed uncertainty quantification as a first-class design principle for distributed intelligence. By integrating representation-aware client assignment and soft-nearest threshold aggregation into a unified federated conformal prediction framework, TrustFed demonstrates that privacy, reliability, and efficiency need not be competing objectives. Instead, they can be jointly achieved through principled, sample-adaptive uncertainty modeling. We anticipate that this paradigm, where privacy-preserving collaboration is paired with rigorous uncertainty guarantees, will play a central role in the next generation of trustworthy machine learning systems, both within healthcare and across safety-critical domains where reliable decision-making under uncertainty is essential.

\section*{Data availability}
The dataset used in this study is publicly available and can be accessed at \href{https://medmnist.com/}{Link}.

\section*{Code availability}
On acceptance, all the source codes to reproduce the results in this study will be made available to the public on GitHub by the corresponding author.

\section*{Acknowledgements}
V. Kumar acknowledges the financial support received from the Ministry of Human Resource Development (MHRD), India in form of the Prime Minister's Research Fellows (PMRF) scholarship. S. Chakraborty acknowledges the financial support received from Anusandhan National Research Foundation (ANRF) via grant no. CRG/2023/007667.

\section*{Author contributions statement}
%V. Kumar: Conceptualization, Methodology, Software, Formal analysis, Writing - original draft. S. Chakraborty: Conceptualization, Methodology, Software debugging, review and editing, Supervision, Funding acquisition. 
VK: Conceptualization, Methodology, Software, Formal analysis, Writing - original draft. SC: Conceptualization, Methodology, Software debugging, review and editing, Supervision, Funding acquisition. SBA: Editing, drafting, and supervision.

\section*{Competing interests} 
The authors declare no competing interests.

%Bibliography
\bibliographystyle{unsrt}  
%\bibliography{references}

%\bibliography{cas-refs}

\begin{thebibliography}{10}

\bibitem{world2025global}
World~Health Organization.
\newblock {\em Global strategy on digital health 2020-2027}.
\newblock World Health Organization, 2025.

\bibitem{organisation2019health}
Organisation for Economic Co-operation and Development (OECD).
\newblock Health in the 21st century: putting data to work for stronger health systems, 2019.

\bibitem{price2019privacy}
W~Nicholson Price and I~Glenn Cohen.
\newblock Privacy in the age of medical big data.
\newblock {\em Nature medicine}, 25(1):37--43, 2019.

\bibitem{rieke2020future}
Nicola Rieke, Jonny Hancox, Wenqi Li, Fausto Milletari, Holger~R Roth, Shadi Albarqouni, Spyridon Bakas, Mathieu~N Galtier, Bennett~A Landman, Klaus Maier-Hein, et~al.
\newblock The future of digital health with federated learning.
\newblock {\em NPJ digital medicine}, 3(1):119, 2020.

\bibitem{edemekong2018health}
Peter~F Edemekong, Pavan Annamaraju, and Micelle~J Haydel.
\newblock Health insurance portability and accountability act, 2018.

\bibitem{regulation2018general}
Protection Regulation.
\newblock General data protection regulation.
\newblock {\em Intouch}, 25:1--5, 2018.

\bibitem{gupta2024digital}
Nidhi Gupta and Ammu George.
\newblock Digital personal data protection act, 2023: Charting the future of india's data regulation.
\newblock In {\em Data Governance and the Digital Economy in Asia}, pages 34--53. Routledge, 2024.

\bibitem{mcmahan2017communication}
Brendan McMahan, Eider Moore, Daniel Ramage, Seth Hampson, and Blaise~Aguera y~Arcas.
\newblock Communication-efficient learning of deep networks from decentralized data.
\newblock In {\em Artificial intelligence and statistics}, pages 1273--1282. PMLR, 2017.

\bibitem{li2020federated}
Tian Li, Anit~Kumar Sahu, Manzil Zaheer, Maziar Sanjabi, Ameet Talwalkar, and Virginia Smith.
\newblock Federated optimization in heterogeneous networks.
\newblock {\em Proceedings of Machine learning and systems}, 2:429--450, 2020.

\bibitem{reddi2020adaptive}
Sashank Reddi, Zachary Charles, Manzil Zaheer, Zachary Garrett, Keith Rush, Jakub Kone{\v{c}}n{\`y}, Sanjiv Kumar, and H~Brendan McMahan.
\newblock Adaptive federated optimization.
\newblock {\em arXiv preprint arXiv:2003.00295}, 2020.

\bibitem{acar2021federated}
Durmus Alp~Emre Acar, Yue Zhao, Ramon~Matas Navarro, Matthew Mattina, Paul~N Whatmough, and Venkatesh Saligrama.
\newblock Federated learning based on dynamic regularization.
\newblock {\em arXiv preprint arXiv:2111.04263}, 2021.

\bibitem{kairouz2021advances}
Peter Kairouz, H~Brendan McMahan, Brendan Avent, Aur{\'e}lien Bellet, Mehdi Bennis, Arjun~Nitin Bhagoji, Kallista Bonawitz, Zachary Charles, Graham Cormode, Rachel Cummings, et~al.
\newblock Advances and open problems in federated learning.
\newblock {\em Foundations and trends in machine learning}, 14(1--2):1--210, 2021.

\bibitem{bonawitz2022federated}
Keith Bonawitz, Peter Kairouz, Brendan McMahan, and Daniel Ramage.
\newblock Federated learning and privacy.
\newblock {\em Communications of the ACM}, 65(4):90--97, 2022.

\bibitem{sheller2020federated}
Micah~J Sheller, Brandon Edwards, G~Anthony Reina, Jason Martin, Sarthak Pati, Aikaterini Kotrotsou, Mikhail Milchenko, Weilin Xu, Daniel Marcus, Rivka~R Colen, et~al.
\newblock Federated learning in medicine: facilitating multi-institutional collaborations without sharing patient data.
\newblock {\em Scientific reports}, 10(1):12598, 2020.

\bibitem{kaissis2020secure}
Georgios~A Kaissis, Marcus~R Makowski, Daniel R{\"u}ckert, and Rickmer~F Braren.
\newblock Secure, privacy-preserving and federated machine learning in medical imaging.
\newblock {\em Nature Machine Intelligence}, 2(6):305--311, 2020.

\bibitem{dayan2021federated}
Ittai Dayan, Holger~R Roth, Aoxiao Zhong, Ahmed Harouni, Amilcare Gentili, Anas~Z Abidin, Andrew Liu, Anthony~Beardsworth Costa, Bradford~J Wood, Chien-Sung Tsai, et~al.
\newblock Federated learning for predicting clinical outcomes in patients with covid-19.
\newblock {\em Nature medicine}, 27(10):1735--1743, 2021.

\bibitem{pati2022federated}
Sarthak Pati, Ujjwal Baid, Brandon Edwards, Micah Sheller, Shih-Han Wang, G~Anthony Reina, Patrick Foley, Alexey Gruzdev, et~al.
\newblock Federated learning enables big data for rare cancer boundary detection.
\newblock {\em Nature Communications}, 13(1):7346, 2022.

\bibitem{kelly2019key}
Christopher~J Kelly, Alan Karthikesalingam, Mustafa Suleyman, Greg Corrado, and Dominic King.
\newblock Key challenges for delivering clinical impact with artificial intelligence.
\newblock {\em BMC medicine}, 17(1):195, 2019.

\bibitem{zhao2018federated}
Yue Zhao, Meng Li, Liangzhen Lai, Naveen Suda, Damon Civin, and Vikas Chandra.
\newblock Federated learning with non-iid data.
\newblock {\em arXiv preprint arXiv:1806.00582}, 2018.

\bibitem{liu2022real}
Fang Liu and Demosthenes Panagiotakos.
\newblock Real-world data: a brief review of the methods, applications, challenges and opportunities.
\newblock {\em BMC Medical Research Methodology}, 22(1):287, 2022.

\bibitem{li2022federated}
Qinbin Li, Yiqun Diao, Quan Chen, and Bingsheng He.
\newblock Federated learning on non-{IID} data silos: An experimental study.
\newblock In {\em IEEE International Conference on Data Engineering}, pages 965--978. IEEE, 2022.

\bibitem{johnson2019survey}
Justin~M Johnson and Taghi~M Khoshgoftaar.
\newblock Survey on deep learning with class imbalance.
\newblock {\em Journal of Big Data}, 6(1):1--54, 2019.

\bibitem{de1980partial}
Bruno De~Finetti.
\newblock Partial exchangeability.
\newblock {\em Studies in inductive logic and probability}, 2:193, 1980.

\bibitem{guo2017calibration}
Chuan Guo, Geoff Pleiss, Yu~Sun, and Kilian~Q Weinberger.
\newblock On calibration of modern neural networks.
\newblock In {\em International conference on machine learning}, pages 1321--1330. PMLR, 2017.

\bibitem{ovadia2019can}
Yaniv Ovadia, Emily Fertig, Jie Ren, Zachary Nado, David Sculley, Sebastian Nowozin, Joshua Dillon, Balaji Lakshminarayanan, and Jasper Snoek.
\newblock Can you trust your model's uncertainty? evaluating predictive uncertainty under dataset shift.
\newblock {\em Advances in neural information processing systems}, 32, 2019.

\bibitem{kompa2021second}
Benjamin Kompa, Jasper Snoek, and Andrew~L Beam.
\newblock Second opinion needed: communicating uncertainty in medical machine learning.
\newblock {\em NPJ Digital Medicine}, 4(1):4, 2021.

\bibitem{marconi2025show}
Luca Marconi and Federico Cabitza.
\newblock Show and tell: A critical review on robustness and uncertainty for a more responsible medical ai.
\newblock {\em International Journal of Medical Informatics}, page 105970, 2025.

\bibitem{loftus2022uncertainty}
Tyler~J Loftus, Benjamin Shickel, Matthew~M Ruppert, Jeremy~A Balch, Tezcan Ozrazgat-Baslanti, Patrick~J Tighe, Philip~A Efron, William~R Hogan, Parisa Rashidi, Gilbert~R Upchurch~Jr, et~al.
\newblock Uncertainty-aware deep learning in healthcare: a scoping review.
\newblock {\em PLOS digital health}, 1(8):e0000085, 2022.

\bibitem{esteva2019guide}
Andre Esteva, Alexandre Robicquet, Bharath Ramsundar, Volodymyr Kuleshov, Mark DePristo, Katherine Chou, Claire Cui, Greg Corrado, Sebastian Thrun, and Jeff Dean.
\newblock A guide to deep learning in healthcare.
\newblock {\em Nature medicine}, 25(1):24--29, 2019.

\bibitem{antunes2022federated}
Rodolfo~Stoffel Antunes, Cristiano Andr{\'e}~da Costa, Arne K{\"u}derle, Imrana~Abdullahi Yari, and Bj{\"o}rn Eskofier.
\newblock Federated learning for healthcare: Systematic review and architecture proposal.
\newblock {\em ACM Transactions on Intelligent Systems and Technology (TIST)}, 13(4):1--23, 2022.

\bibitem{kendall2017uncertainties}
Alex Kendall and Yarin Gal.
\newblock What uncertainties do we need in bayesian deep learning for computer vision?
\newblock {\em Advances in neural information processing systems}, 30, 2017.

\bibitem{lakshminarayanan2017simple}
Balaji Lakshminarayanan, Alexander Pritzel, and Charles Blundell.
\newblock Simple and scalable predictive uncertainty estimation using deep ensembles.
\newblock {\em Advances in neural information processing systems}, 30, 2017.

\bibitem{vovk2005algorithmic}
Vladimir Vovk, Alexander Gammerman, and Glenn Shafer.
\newblock {\em Algorithmic learning in a random world}, volume~29.
\newblock Springer, 2005.

\bibitem{begoli2019need}
Edmon Begoli, Tanmoy Bhattacharya, and Dimitri Kusnezov.
\newblock The need for uncertainty quantification in machine-assisted medical decision making.
\newblock {\em Nature Machine Intelligence}, 1(1):20--23, 2019.

\bibitem{lu2023federated}
Charles Lu, Yaodong Yu, Sai~Praneeth Karimireddy, Michael Jordan, and Ramesh Raskar.
\newblock Federated conformal predictors for distributed uncertainty quantification.
\newblock In {\em International Conference on Machine Learning}, pages 22942--22964. PMLR, 2023.

\bibitem{plassier2023conformal}
Vincent Plassier, Mehdi Makni, Aleksandr Rubashevskii, Eric Moulines, and Maxim Panov.
\newblock Conformal prediction for federated uncertainty quantification under label shift.
\newblock In {\em International Conference on Machine Learning}, pages 27907--27947. PMLR, 2023.

\bibitem{angelopoulos2023conformal}
Anastasios~N Angelopoulos, Stephen Bates, et~al.
\newblock Conformal prediction: A gentle introduction.
\newblock {\em Foundations and trends in machine learning}, 16(4):494--591, 2023.

\bibitem{kumar2026cortinetphysicsperceptionhybridcorticalinspired}
Vagish Kumar and Souvik Chakraborty.
\newblock Cortinet: A physics-perception hybrid cortical-inspired dual-stream network for gallbladder disease diagnosis from ultrasound, 2026.

\bibitem{yang2023medmnist}
Jiancheng Yang, Rui Shi, Donglai Wei, Zequan Liu, Lin Zhao, Bilian Ke, Hanspeter Pfister, and Bingbing Ni.
\newblock Medmnist v2-a large-scale lightweight benchmark for 2d and 3d biomedical image classification.
\newblock {\em Scientific Data}, 10(1):41, 2023.

\bibitem{geyer2017differentially}
Robin~C Geyer, Tassilo Klein, and Moin Nabi.
\newblock Differentially private federated learning: A client level perspective.
\newblock {\em arXiv preprint arXiv:1712.07557}, 2017.

\end{thebibliography}

\end{document}